\documentclass[11pt,journal,compsoc]{IEEEtran}
\usepackage{amsmath}
\usepackage{url}
\usepackage{graphicx}
\usepackage{tabularx}
\usepackage{makecell}

\usepackage{physics}
\usepackage{amsmath}
\usepackage{tikz}
\usepackage{mathdots}
\usepackage{yhmath}
\usepackage{cancel}
\usepackage{color}
\usepackage{siunitx}
\usepackage{array}
\usepackage{multirow}
\usepackage{amssymb}
\usepackage{gensymb}
\usepackage{tabularx}
\usepackage{booktabs}
\usetikzlibrary{fadings}
\usetikzlibrary{patterns}
\usetikzlibrary{shadows.blur}
\usetikzlibrary{shapes}

\begin{document}
\title{Gait Design of a Novel Arboreal Concertina Locomotion for Snake-like Robots}

\author{Shuoqi~Chen, Aaron~M.~Roth \\ Carnegie Mellon University}

\IEEEtitleabstractindextext{%
\begin{abstract}
In this paper, we propose a novel strategy for a snake robot to move straight up a cylindrical surface. Prior works on pole-climbing for a snake robot mainly utilized a rolling helix gait, and although proven to be efficient, it does not reassemble movements made by a natural snake. We take inspiration from nature and seek to imitate the Arboreal Concertina Locomotion (ACL) from real-life serpents. In order to represent the 3D curves that make up the key motion patterns of ACL, we establish a set of parametric equations that identify periodic functions, which produce a sequence of backbone curves. We then build up the gait equation using the curvature integration method, and finally, we propose a simple motion estimation strategy using virtual chassis and non-slip model assumptions. We present experimental results using a 20-DOF snake robot traversing outside of a straight pipe.
\end{abstract}
}

\maketitle
\IEEEdisplaynontitleabstractindextext
\IEEEpeerreviewmaketitle
\section{Introduction and Related Work}\label{sec:introduction}


Arboreal Concertina Locomotion (ACL) is a common locomotion strategy used by snakes in nature to climb up bare branches or smooth poles. \cite{jayne2020defines} In this mode of motion, snakes climb by gripping or anchoring with portions of their body while pulling or pushing other sections in the direction of movement. In other words, snakes form themselves in the shape of a pitch-varying helix and inch upward in a periodic fashion. Snake-like robots could benefit from mimicking this motion pattern and learning to climb up cylindrical objects efficiently. However, only few previous literature offers a mathematical model for describing the motion pattern of ACL~\cite{tang2017arboreal}, and even fewer research groups have attempted to implement the gait on real snake robots. Instead, the most commonly used pole climbing strategy was the rolling helix gait~\cite{zhen2015modeling}.

In this paper, we propose a novel gait design that approximates the ACL adopted by natural snakes. We build upon the results of ~\cite{hatton2010generating} (which introduces a keyframe algorithm), and add in a discretization of the backbone curve. To facilitate the motion modeling of ACL, we take insights from~\cite{rollinson2011virtual} and make more accurate modeling of the contact points.

\section{Model}
\begin{figure}[!]
\centering

\includegraphics[width=0.2\textwidth]{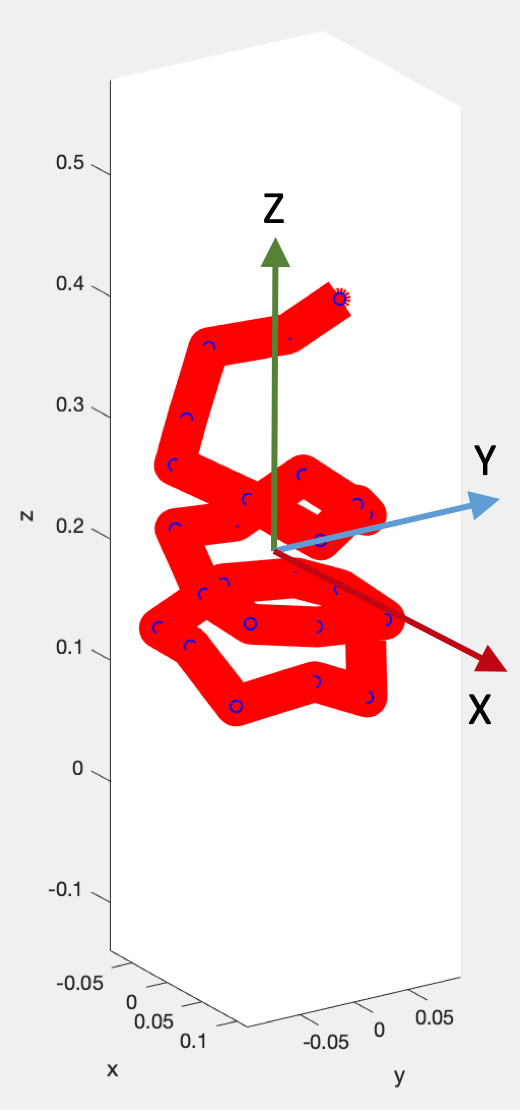} 

\caption{A virtual chassis visualization of a snake robot executing the designed Arboreal Concertina Locomotion (ACL) gait. The direction of motion is along the $z$ axis.}
\label{fig:virtual}
\end{figure}

\subsection{Parametric Equations for the backbone curve}

A natural snake alternatively contracts and expands its body as it moves up straight poles. This motion is similar to that of the longitudinal wave, where the displacement of the section of the wave body is along the direction of the propagation. This motion pattern can be captured by the general parametric equations of a pitch-varying helix as shown in equation \ref{eqn:parametric1}. 

\begin{equation}\label{eqn:parametric1}
\begin{split}
x =& \cos(t)\cdot (r + A_r \cos(k_r t + \omega h )) \\
y =& \sin(t)\cdot (r + A_r \cos(k_r t + \omega h )) \\
z =& pt + A_p \sin (\Omega t + \omega h)
\end{split}
\end{equation}

where $r$ denotes the radius, and $p$ is the pitch of the base helix model, $t$ is the sequence of time and $h$ is the non-curve-length parametrization of a 3D curve. $A_r$ and $A_p$ are coefficients describing the magnitude of lateral and longitudinal extension and contraction, and $k_r$, $\Omega$ and $\omega$ determine the frequency of the aforementioned expansion and contraction.

Notice that for snake robots that remain in contact with the pole at all times during the gait cycle, the parametric equations can be simplified to the form described in equation \ref{eqn:parametric2}.

\begin{equation}\label{eqn:parametric2}
\begin{split}
x =& r \cdot \cos(t) \\
y =& r \cdot \sin(t) \\
z =& pt + A_p \sin(\Omega t + \omega h)
\end{split}
\end{equation}

Setting $A_r$ = 0 simplifies calculation without losing the generality of the motion model. We will use the simplified parametric equations for the rest of the paper. It is important to note that equation \ref{eqn:parametric2} has a form similar to that of a constant helix with an addition of a compounded sinusoidal function. Here, $\Omega$ describes the shape of the pitch-varying helix at a fixed time point, whereas $\omega$ describes how this shape changes over time $t \in [0, T]$.

\subsection{Gait Equations}

Since the snake robots are a system of floating base, hyper-redundant manipulators, it is difficult to generate a full dynamic model. Therefore, we choose to manipulate the snake robot on the kinematic level and use joint angles as position commands as the form of control.

To calculate the set of joint angles over the period of an ACL gait cycle, we first discretize the continuous motion into key frame sequences and parameterize the 3D curve of the pitch-varying helix, also known as the backbone curves, that produce those sequences. Then, we use the curvature integration algorithm~\cite{yamada2006study} to calculate the set of joint angles that fit the snake robot to the backbone curves.

Since the modular snake robot is composed of multiple identical modules whose joints alternatively rotate in orthogonal directions, the curvature integration algorithm we use employs the bellows model~\cite{yamada2006study} representing the shape of a backbone curve using both the dorsal curvature, $\displaystyle \kappa _{d}( s)$ and lateral curvature, $\displaystyle \kappa _{l}( s)$, where their relationships with the corresponding dorsal and lateral joint angles are shown in equation \ref{eqn:doral_lateral}.

\begin{equation}\label{eqn:doral_lateral}
\begin{split}
\alpha _{dorsal}( i) \ =\ \int ^{( i+1) l}_{( i-1) l} \kappa _{d}( s) \ ds\\
\\ 
\alpha _{lateral}( i) \ =\ \int ^{( i+1) l}_{( i-1) l} \kappa _{l}( s) \ ds
\end{split}    
\end{equation}

where index $i$ corresponds to the sequences of dorsal and lateral joints, respectively, and $l$ is the module length.

Additionally, we can instill temporal change of the shape into the joint angles to include expansion and contraction, which allow the "inching" motion. The complete forms of equations to calculate the dorsal and lateral joint angles are shown in equation \ref{eqn:gait1} and \ref{eqn:gait2}, where $s$ is the curve length to be integrated, $\phi_0$ is the initial angle offset, and $\kappa$ and $\tau$ are curvature and torsion of the backbone curve at each keyframe. It is difficult to derive a closed-form solution for the ACL gait. Therefore, we numerically calculate the curvature, torsion, and subsequently, the joint angles in our experiments. 

\begin{figure*}[!h]
\normalsize
\begin{equation}\label{eqn:gait1}
\alpha _{lateral} (i, t) = 
\sin(\phi_0 (t)) \int_{l\cdot(i-1)}^{l\cdot(i+1)} \left( \kappa(s)\cdot \cos \left( \int_0^s \tau(s)ds \right) \right) ds + 
\cos(\phi_0 (t)) \int_{l\cdot(i-1)}^{l\cdot(i+1)} \left( \kappa(s)\cdot \sin \left( \int_0^s \tau(s)ds \right) \right) ds
\end{equation}
\end{figure*}

\begin{figure*}[!h]
\normalsize
\begin{equation}\label{eqn:gait2}
\alpha _{dorsal}(i, t) = 
\cos(\phi_0 (t)) \int_{l\cdot(i-1)}^{l\cdot(i+1)} \left( \kappa(s)\cdot \cos \left( \int_0^s \tau(s)ds \right) \right) ds -
\sin(\phi_0 (t)) \int_{l\cdot(i-1)}^{l\cdot(i+1)} \left( \kappa(s)\cdot \sin \left( \int_0^s \tau(s)ds \right) \right) ds
\end{equation}
\end{figure*}

\subsection{Motion Estimation}

Representing the motion of snake robots in the world frame requires additional knowledge of the contact model, because the internal shape changes that the robot uses to interact with the world are quite complex. To mitigate this problem, we propose a simple strategy to estimate the displacement of the snake robot with respect to the world frame by employing a virtual chassis and three simplifying assumptions. The virtual chassis defines a body frame such that the position of the robot is at the center of mass (COM) of the current robot configuration, and orientation can be estimated by performing principal component analysis on zero-mean data.~\cite{enner2013motion} 

To locate the geometric center of a given snake robot configuration at a fixed time step, we take the mean positions of all module COMs and define the result as the body frame origin. Next, we find the orientation of the virtual chassis by aligning the principal axes of the chassis body frame with the principal moments of inertia of the virtual chassis ($I_{vc}$) of the modules. The orientation vectors are the columns of V after performing the singular value decomposition on the position difference matrix P.
\begin{equation}
\boldsymbol{P} \ =\ \left[ x\ -\ \overline{x} ,\ \ y\ -\ \overline{y} ,\ z\ -\ \overline{z}\right]    
\end{equation}
\begin{equation}
USV^{T} \ =\ \boldsymbol{P}
\end{equation}

Upon defining the virtual chassis for each time frame, we iteratively search for contact points between the snake robot and the pole and then estimate the displacement by compensating the COM shift between two consecutive frames. We make three simplifying assumptions to guarantee the existence of unique contact points at each time step: 1) The snake robot makes non-slip contact with its environment, 2) The contact points always happen around the joints, and 3) for every two-link segment on the snake robot, there must be at least one contact point whose euclidean distance to the pole center line is the smallest. Following the assumptions, the overall translational displacement of the snake robot over one ACL gait cycle can be estimated in equation \ref{eqn:disp_calc}.

\begin{equation}\label{eqn:disp_calc}
\begin{split}
    X_{rigid\_body} = \left( \sum_{t = 0}^{T}{\delta x} \right) + 2\pi r_{joint} \\
    \delta x = - \left( \frac{\sum_{i=1}^{k}{x_k^{t+1}}}{k} - \frac{\sum_{i=1}^{k}{x_k^{t}}}{k} \right)
\end{split}
\end{equation}

where k is the number of joints in contact with the cylinder, $x_k$ is the axial directional component of the corresponding contact joints. 
In the equation for $\delta x$, $t$ and $t+1$ are the current time steps where the contact joints are identified. 

\begin{figure*}[!]
\centering
\resizebox{1\textwidth}{!}{
\includegraphics[height=5cm]{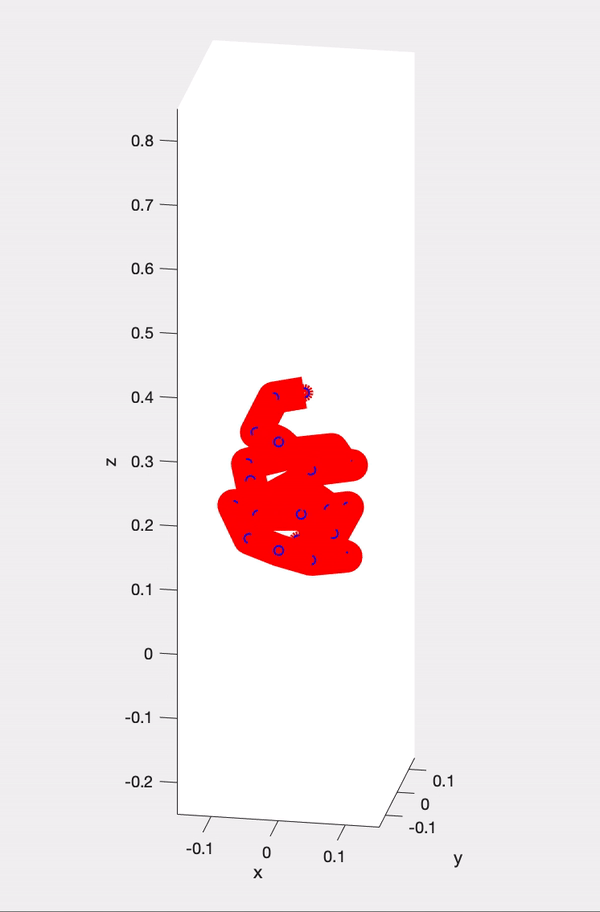} 
\includegraphics[height=5cm]{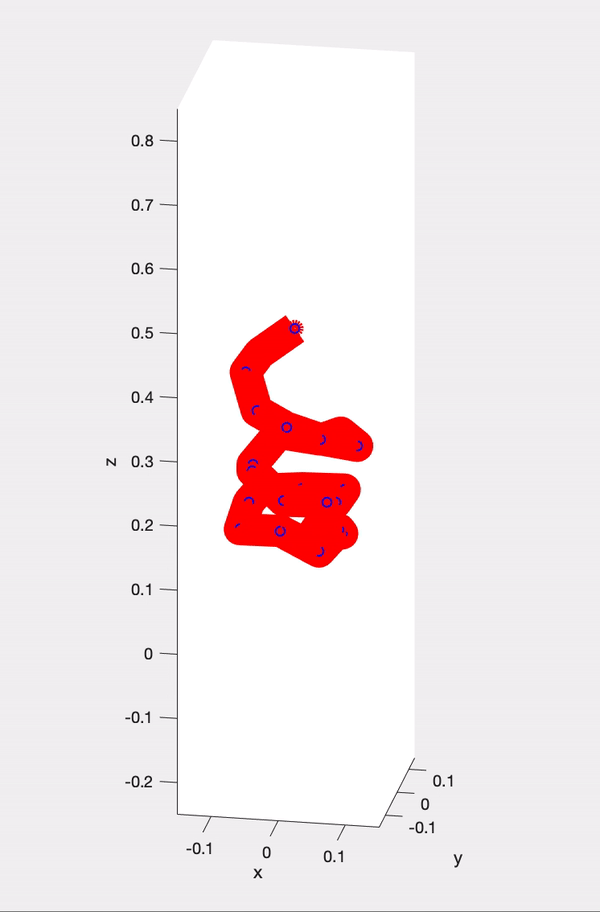} 
\includegraphics[height=5cm]{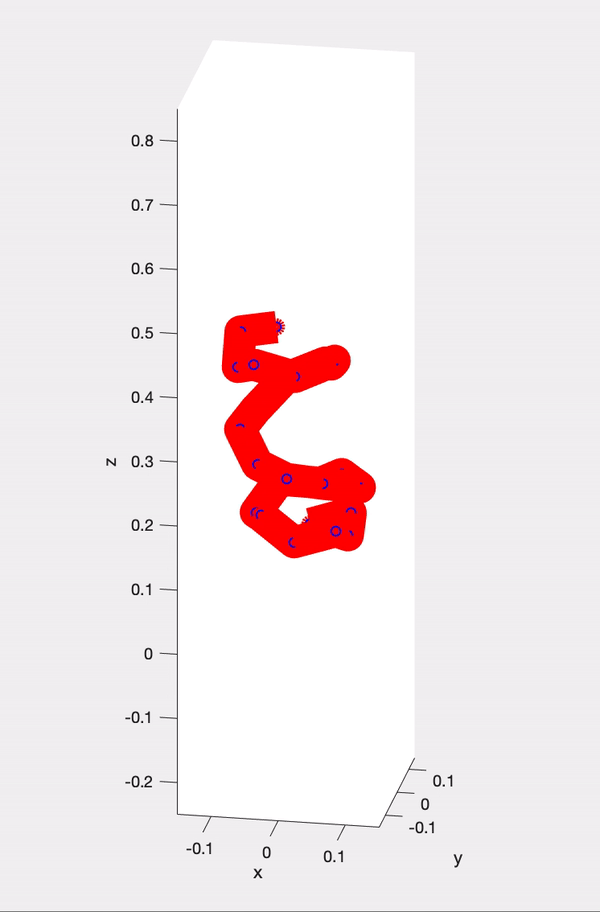} 
\includegraphics[height=5cm]{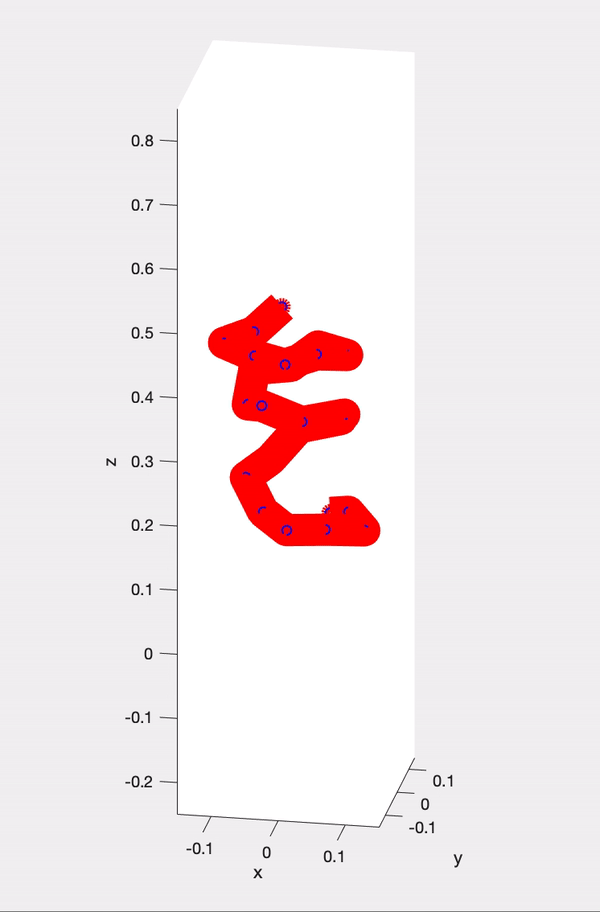} 
\includegraphics[height=5cm]{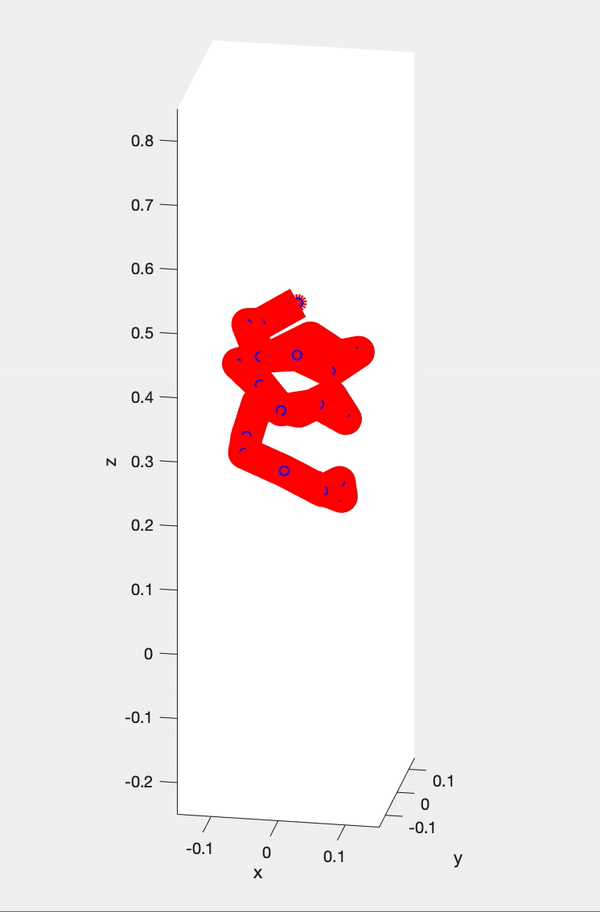}
\includegraphics[height=5cm]{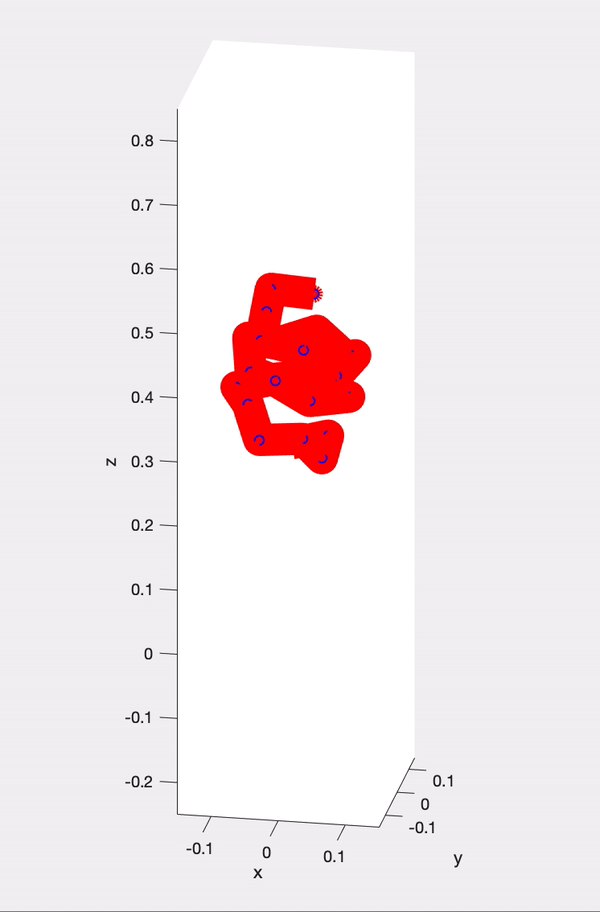}
}
\label{fig:sim}
\end{figure*}

\begin{figure*}[!]
\centering
\resizebox{0.98\textwidth}{!}{
\includegraphics[height=3cm]{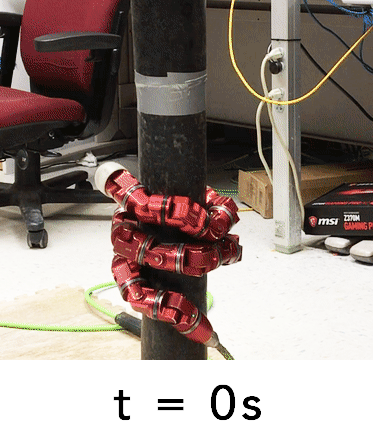} 
\includegraphics[height=3cm]{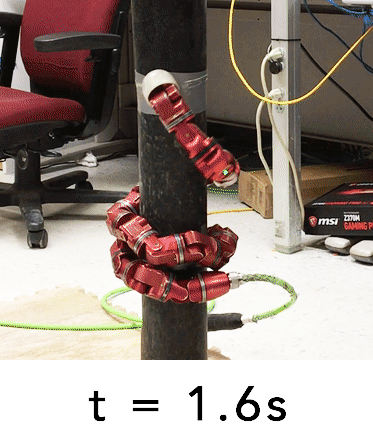} 
\includegraphics[height=3cm]{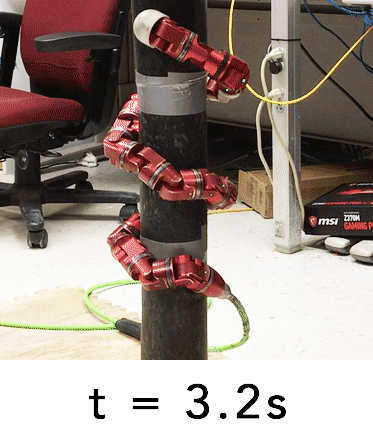} 
\includegraphics[height=3cm]{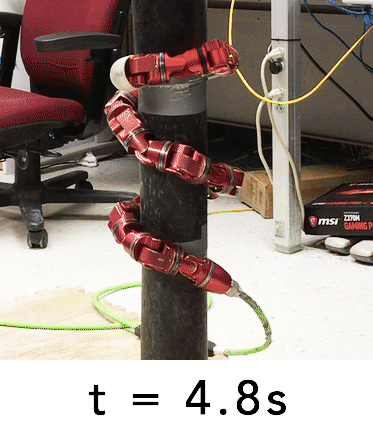} 
\includegraphics[height=3cm]{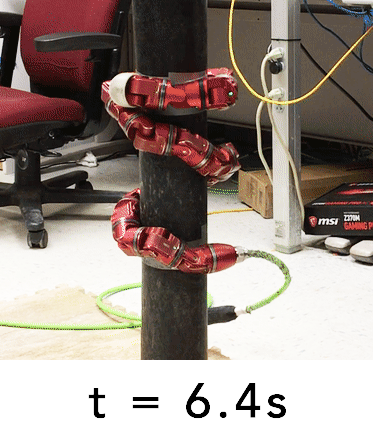}
\includegraphics[height=3cm]{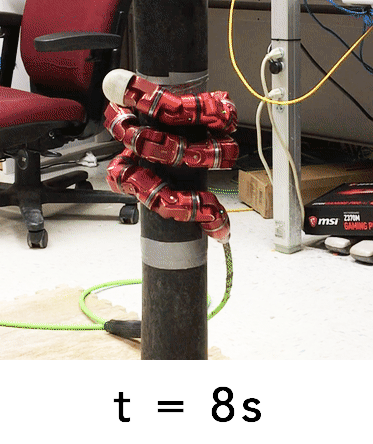}
}
\caption{Top row: simulated locomotion pattern of a snake robot of 20 articulated joints using the ACI gait. Bottom row: pole climbing experiment of a 20 DOF SEA snake robot using the aforementioned gait.}
\label{fig:exp}
\end{figure*}

\section{Results}

We retrieved the displacements per ACL gait cycle in both simulation and robot experiments in 8 different parameter settings. The robot we used is the 20-DOF series elastic actuator (SEA) snake robot. We discretized one gait cycle to 220 time steps and passed in the joint angles as position commands at 10 Hz frequency. The parameters and the displacement result of the ACL gait are summarized in table \ref{tab:results_table}.

\begin{table}[]
    \centering
    \resizebox{0.5\textwidth}{!}{
    \begin{tabular}{r|c|c|c|c|c|c|c|c|c}
         \textbf{Trial No.} & 1 & 2 & 3 & 4 & 5 & 6 & 7 & 8 \\ 
         \hline
         \makecell{\textbf{Estimated} \\ \textbf{Disp. (cm)}}
          & 46.22 & 46.54 & 46.60 & 48.00 & 47.94 & 50.08 & 48.98 & 50.17 \\
          \hline
         \makecell{\textbf{Actual} \\ \textbf{Disp. (cm)}} & 52.99 & 47.93 & 49.63 & 49.63 & 49.63 & 47.10 & 48.79 & 48.79 \\
         \hline
         \textbf{Error \%} & 12.8 & 0.8 & 6.5 & 3.3 & 3.5 & 6.0 & 0.4 & 2.8  \\
    \end{tabular}
    }
    \caption{Comparison of the simulated and the measured Displacements using the ACL gait over two cycles in 8 different experiments on a 20 DOF SEA snake robot.}
    \label{tab:results_table}
\end{table}

The estimated displacements are close to the experimentally measured displacement. Even though the assumptions we make exclude dynamic information such as frictional forces and intermittent contact, our motion estimation strategy still provides reasonable displacement estimates at little computational expense compared to full dynamic simulation in state-of-the-art software.

We also compared the performances of the well-established rolling helix gait and the ACL gait in terms of distance traveled per cycle in experiments. Table \ref{tab:results_table_acl_vs_rolling} shows the results.

\begin{table}[]
    \centering
    \begin{tabular}{c|c|c}
          & \textbf{Rolling Helix Gait} & \textbf{ACL Gait} \\
         \hline
         \textbf{Mean (cm)} & 47.09 & 49.06 \\
        \hline
        \makecell{\textbf{Standard} \\ \textbf{Deviation (cm)}} & 0.99 & 1.78 \\
    \end{tabular}
    \caption{Displacement comparison between the rolling helix gait and the ACL gait over 2 cycles executed on a 20 DOF SEA snake robot.}
    \label{tab:results_table_acl_vs_rolling}
\end{table}

\section{Discussion and Conclusion}

Experimental measurements above show that ACL gait has a small increase in displacement per gait cycle compared to the rolling helix gait. Although even a small increase is a success, we had hoped for and expected a larger improvement, and investigated as to why this was not the case. From our observation, the mitigation of the improvement is primarily due to frequent downward slipping during ACL's expanding and contracting phase. This decreases the efficacy of the motion. This issue can be addressed by adding robot force control in parallel with position control, and in addition, surface friction can be increased to reduce slipping by either adding a rubber skin to the snake robot itself or a rough surface coating to the pole.

As the ACL gait proves to be a successful motion pattern for snake robots on straight poles, it provides new insights into designing new pole-climbing gaits and even opens up the possibility for hardware development. For example, one of the difficulties of strapping sensors to the SEA snake module is that, as the module rotates, they make unpredictable contacts with hard surfaces, and in doing so, risk breaking the mounted sensors. In the future, we can potentially revise the ACL gait so that it does not require rolling motion. Such studies could potentially generate pole-climbing gaits with fixed locations on the modules where contacts never happen and therefore additional sensors could be mounted.


\end{document}